\documentclass[conference]{IEEEtran}
\IEEEoverridecommandlockouts
\usepackage{subfig}
\usepackage{url}
\usepackage{cite}
\usepackage{amsmath,amssymb,amsfonts}
\usepackage{algorithmic}
\usepackage{graphicx}
\usepackage{textcomp}
\usepackage{xcolor}
\usepackage{import}
\usepackage{multicol}
\usepackage{lipsum}
\usepackage{mwe}
\usepackage{pifont}
\usepackage{caption}
\usepackage{booktabs}
\usepackage{hyperref}

\usepackage{eso-pic}
\newcommand\AtPageUpperMyright[1]{\AtPageUpperLeft{%
 \put(\LenToUnit{0.5\paperwidth},\LenToUnit{-1cm}){%
     \parbox{0.5\textwidth}{\raggedleft\fontsize{9}{11}\selectfont #1}}%
 }}%
\newcommand{\conf}[1]{%
\AddToShipoutPictureBG*{%
\AtPageUpperMyright{#1}
}
}

\DeclareCaptionLabelFormat{cont}{#1~#2\alph{ContinuedFloat}}
\def\BibTeX{{\rm B\kern-.05em{\sc i\kern-.025em b}\kern-.08em
    T\kern-.1667em\lower.7ex\hbox{E}\kern-.125emX}}

\begin{document}

\title{Real Time Bangladeshi Sign Language Detection using Faster R-CNN
}

\author{
\IEEEauthorblockN{Oishee Bintey Hoque\IEEEauthorrefmark{1},
Mohammad Imrul Jubair\IEEEauthorrefmark{2},
Md. Saiful Islam\IEEEauthorrefmark{3},
Al-Farabi Akash\IEEEauthorrefmark{4},
Alvin Sachie Paulson\IEEEauthorrefmark{5}
}
\IEEEauthorblockA{
Department of Computer Science and Engineering,\\
Ahsanullah University of Science and Technology
Dhaka, Bangladesh\\
\{\IEEEauthorrefmark{1}bintu3003,
\IEEEauthorrefmark{3}saiful.somum,
\IEEEauthorrefmark{4}alfa.farabi,
\IEEEauthorrefmark{5}sachiekhan\}@gmail.com,
\IEEEauthorrefmark{2}mohammadimrul.jubair@ucalgary.ca
}
}
\IEEEoverridecommandlockouts
\IEEEpubid{}
\maketitle
\IEEEpubidadjcol 

\conf{}
\begin{abstract}
 Bangladeshi Sign Language (BdSL) is a commonly used medium of communication for the hearing-impaired people in Bangladesh. Developing a real time system to detect these signs from images is a great challenge. In this paper, we present a technique to detect BdSL from images that performs in real time. Our method uses Convolutional Neural Network based object detection technique to detect the presence of signs in the image region and to recognize its class. For this purpose, we adopted Faster Region-based Convolutional Network approach and developed a dataset -- \textit{BdSLImset} -- to train our system. Previous research works in detecting BdSL generally depend on external devices while most of the other vision-based techniques do not perform efficiently in real time. Our approach, however, is free from such limitations and the experimental results demonstrate that the proposed method successfully identifies and recognizes Bangladeshi signs in real time.

\end{abstract}


\section{Introduction}
\begin{figure}[htbp]
\centerline{\includegraphics[scale=0.18]{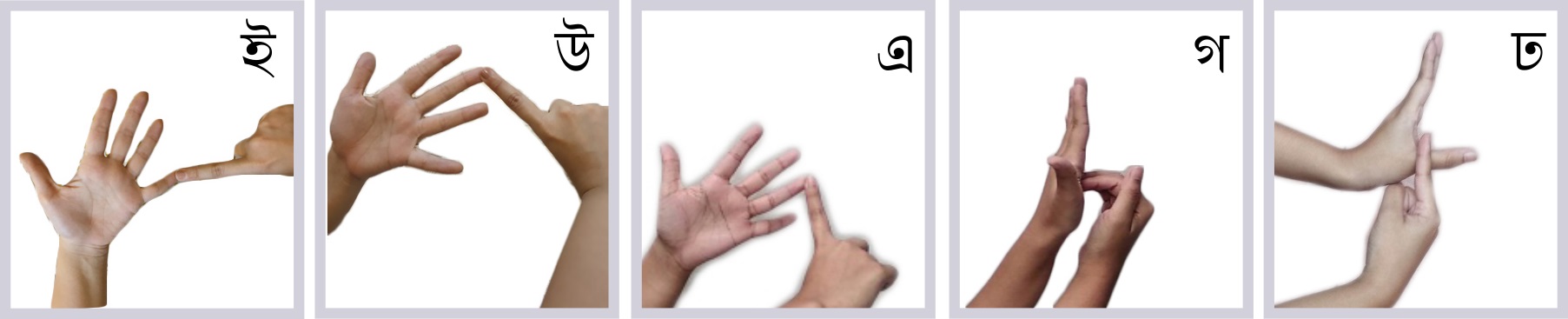}}
\caption{Examples of Sign Language. Here different signs from BdSL are shown.}
\label{fig1}
\end{figure}

The field of computer vision is reaching every possible sectors to help human being. In recent days, it is being used to assist the deaf community by facilitating the \textit{sign language detection technique}. In order to liaise with other people, deaf persons widely use Sign languages as the mediums of communication (Fig. \ref{fig1}). Hence, the main contribution of the sign language detection technique is to act as a digital interpreter between the deaf and the hearing people. There are detection strategies where supplementary equipment such as specialized gloves \cite{b1}, Kinect \cite{b2} etc. are used, however, the inputs are not 2D images and the system is much complicated. Most of the other techniques take image (or sequence of images) containing signs as input using camera, and the ultimate step is to detect those signs and present them in a meaningful manner \cite{b3}. The later approaches do not depend on any additional devices and the image processing techniques are applied on the input image which locates the position of the gesture for detection. Previous works were solely based on old-fashioned image pre-processing methods such as -- morphological operations, color-based foreground segmentation etc. In this era of machine learning, Convolutional neural network provides us more powerful tools for object detection which has surpassed the former approaches by all means. In our work, we explore the area of sign language detection and develop a technique to detect signs in real time by exploiting the \textit{Faster Region-based Convolutional Network} method (faster R-CNN) \cite{b4}.
Different sign languages are used worldwide, such as -- American Sign Language (ASL), Chinese Sign Language (CSL), Parisian Sign Language, etc. ASL recognition has been explored since around $1995$ \cite{b5} and other languages are also investigated by different researchers \cite{b6}\cite{b7}. However, a very few work has been done on Bangladeshi Sign Languages (BdSL). Most of the previous works does not take the advantages of convolutional neural network based object detection technique to identify the gestures. In this paper, we investigate BdSL recognition and develop a system using faster R-CNN. For this purpose, we collected images of specific signs with different backgrounds, variations and built a dataset -- \textit{\textbf{BdSLImset}} (\textit{\textbf{B}angla\textbf{d}eshi \textbf{S}ign \textbf{L}anguage \textbf{Im}age Data\textbf{set}}) -- to train our system. In that case, we do not need to pre-process the images to differentiate the foreground. Our results exhibit satisfactory outcome in identifying the gesture area and recognizing BdSL in real-time.

In summary, the contributions of our work are listed below.
\begin{itemize}
\item We generated a dataset called \textit{\textbf{BdSLImset}} containing images of Bangladeshi signs with random backgrounds and lighting conditions. The dataset is available here: \texttt{\url{https://github.com/imruljubair/bdslimset}}.
\item We propose a recognition technique to identify the signs and detect BdSL from images in real time. Our system is trained on our \textit{\textbf{BdSLImset}} dataset and uses \textit{Faster R-CNN} object detection approach. We have demonstrated the experimental outcomes of our proposed methodology.
\end{itemize}

The remainder of the paper is organized as follows. We reviewed on previous BdSL recognition methods and datasets followed by an overview of Faster R-CNN in Section \ref{BackRel}. We present our dataset and pipeline of the proposed technique in Section \ref{prop}. Section \ref{exp} illustrates our experimental results and comparisons with previous works. Possible avenues of future exploration are discussed in Section \ref{conc}.

\section{Background and Related Works}
\label{BackRel}
In the first part of this section, we discuss about previous works on the domain of Bangladeshi sign language detection followed by a study on the existing datasets. The last section provides a brief review on the Faster R-CNN \cite{b4} which is used in our proposed methodology.

\subsection{Existing BdSL Detection Techniques}
             Rahman, Fatema and Rokonuzzaman used gloves containing dots at each finger position to track the action of signs. The authors collected the dots and mapped the results of the clustered dots to predefined charts. The system can only detect the sign of Bengali numerals from $1$ to $10$ \cite{b8}. In \cite{b9}, researcher applied image processing operations on input images with any assistance of gloves. They determined relative finger tip positions from image and trained an artificial neural network using those tip-position vectors. In their case, the recognition was not in real time and the authors claimed to have an accuracy of $98.99$ percent for Bangla sign letters. In \cite{b10}, the authors introduced a computer vision-based Bengali sign words recognition system which used contour analysis and \textit{Haar-like} feature based cascaded classifier. They trained their classifier and tested the system using $3600$ ($36 \times 10 \times 10$) contour templates for $36$ Bengali sign letter separately and achieved $96.46$ percent recognition accuracy. In $2017$, Yasir, Prasad, Alsadoon and Sreedharan made use of virtual reality by applying leap motion controller to capture hand gestures and implemented CNN for detecting the signs \cite{b1} . However, their recognition was not in real time. In \cite{b17}, the authors also introduced an approach for detecting BdSL letters and digits which applies a fuzzy-logic based model and grid-pattern analysis in real-time.
             In \cite{b18}, the authors presented a real-time Bengali and Chinese numeral signs recognition system using contour matching. The system is trained and tested using a total of $2000$ contour templates separately for both Bengali and Chinese numeral signs from $ 10$ signers and achieved recognition accuracy of $95.80$\%  and $95.90$\%  with computational cost of $8.023$ milliseconds per frame.
             In \cite{b19}, the authors introduced a method of recognizing Hand-Sign-Spelled Bangla language. The system is divided into two phase -- hand sign classification and automatic recognition of hand-sign-spelled for BdSL using Bangla Language Modeling Algorithm (BLMA). The system is tested for BLMA using words, composite numerals and sentences in BdSL achieving mean accuracy of $93.50$\%, $95.50$\% and $90.50$\% respectively.

The existing BdSL recognition techniques are not trained with same datasets. Different models are trained on author's individual datasets which were not available for further exploration. We're going to briefly discuss about the existing BdSL datasets in the next subsection.

\subsection{Existing BdSL Datasets}
Most of the existing datasets used in BdSL detection models do not have variation in background and different lighting condition. Because of these limitations, those datasets are not suitable to be fed into the CNN model. A comparison between others datasets which were used in previous works and our dataset is presented in Table {\ref{tab2}} in later section.

               \begin{figure}[ht]
                \centerline{\includegraphics[scale=0.15]{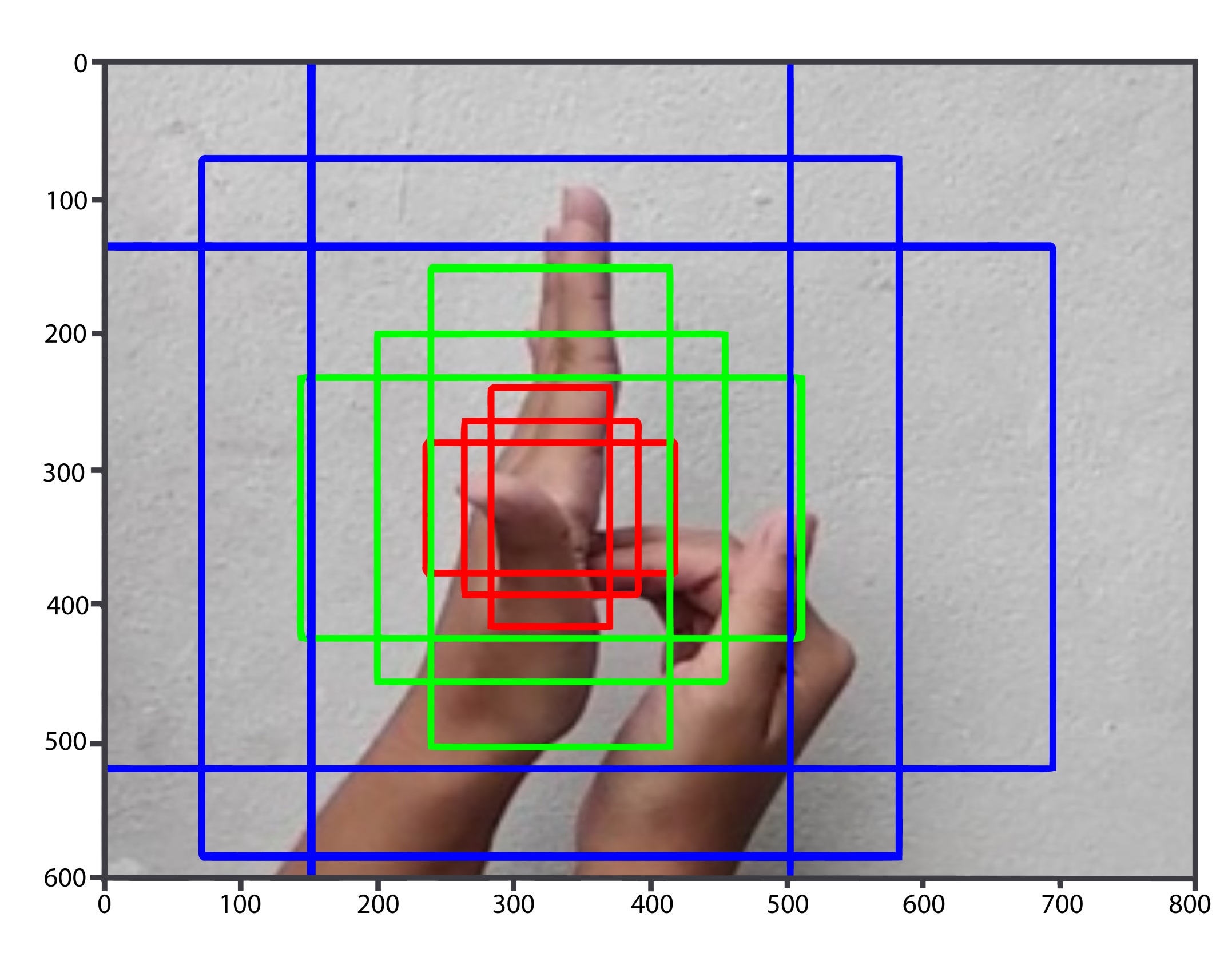}}
                \caption{A conceptual illustration of anchors in CNN feature map for $3$ different aspect ratios and sizes. Here, three colors represent three scales or sizes: $128\times 128$, $256\times 256$, $512 \times512$,
and three boxes have height width ratios $1:1$, $1:2$ and $2:1$ respectively.}
                \label{fig2}
                \end{figure}

\subsection{Faster-RCNN model}
In our work, we emphasize on identifying signs in the image. This challenge basically falls under the domain of object detection. Our study on CNN based object detection leads us to focus on using Faster R-CNN. Faster R-CNN is an improved version of its antecedent algorithms -- R-CNN\cite{b13} and Fast R-CNN\cite{b14}. Unlike other models, Faster R-CNN feeds only necessary region to the convolutional neural network. Initially, CNN generates a feature map and a network -- Regional Proposal Network (RPN) -- proposes regions with high probability of containing desired object. In the network architecture, it applies a \textit{region of interest (RoI)} pooling layer and reshape them into a fixed size to feed into a fully connected layer. From the RoI feature vector, it uses a softmax layer to predict the class of the proposed region.                 \begin{figure*}[ht]
    \centerline{\includegraphics[scale=0.6]{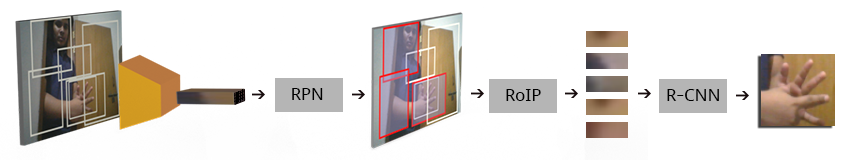}}
    \caption{Detailed architecture of the network. Firstly the input image goes into CNN framework and creates a feature map. RPN proposes anchors with higher probability of being an anchor and the RoI pooling classification is performed at last to finalize the detection.}
    \label{fig8}
    \end{figure*}
A key factor that plays an important role in Faster R-CNN is the \textit{Anchor}. Anchors are fixed sized bounding box. An input image is divided into several anchors in CNN feature map. In proposed model, there are $4$ different scale  ($0.25, 0.5, 1.0, 2.0$)  and $3$ aspect ratios ($0.5, 1.0, 2.0$) with height and width stride of $16$ anchors. Fig. \ref{fig2} shows an example illustration of anchors. Therefore, if an image has $w\times h$ ratio and we choose every stride of $16$, there will be $(w/16) \times (h/16)$ positions to consider and it will finally have $(w/16) \times (h/16)\times 12$ anchors for each image.
The RPN takes all the reference boxes from the feature map and generate a good set of proposals of being objects. 
The process votes the anchor if it is an object or not. The Anchor is labelled as foreground if it is voted as object, otherwise labelled as background. Later, the anchors based on the similar criteria are selected and refined. Anchors labelled as background is not included in the regression.

After the RPN stage, we get proposed regions with different sizes. RoI pooling reduce the feature maps into the same size. It eventually splits the input feature map into a fixed number of roughly equal regions then applies Max-Pooling on every region. Hence the output of RoI Pooling is always fixed regardless the size of input.

\begin{figure*}[!ht]
\centerline{\includegraphics[scale=0.4]{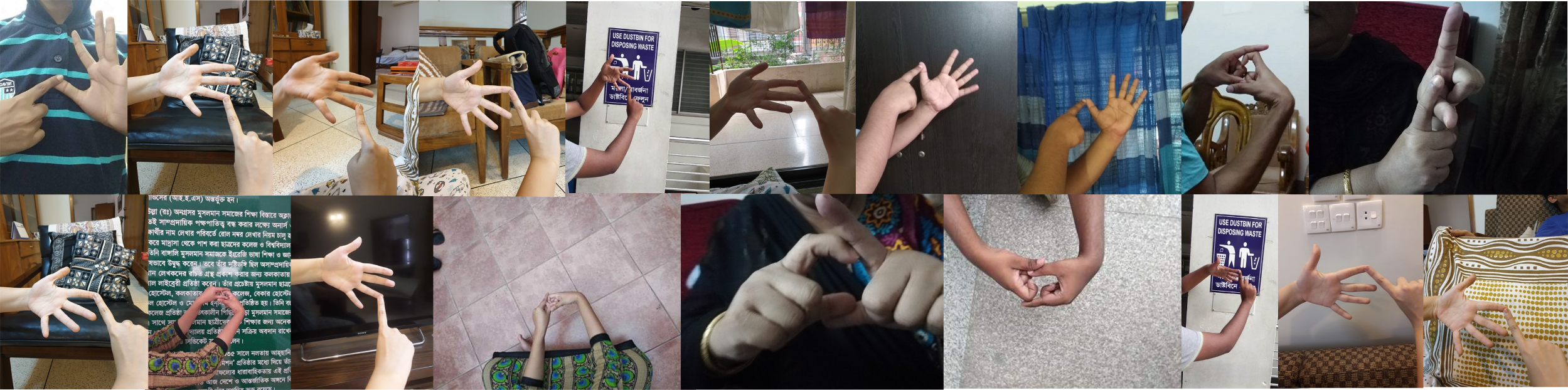}}
\caption{Sample sign language images from our \textit{BdSLImset} dataset for different Bengali letters with different background.}
\label{fig4}
\end{figure*}
\begin{figure}[htb]
\centerline{\includegraphics[scale=1.0]{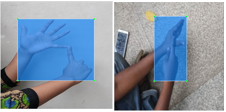}}
\caption{Images with labels of different signs (labelled inside the blue boxes).}
\label{fig5}
\end{figure}

\section{Proposed Methodology}
\label{prop}
In this section, we first present our dataset followed by the illustration of our proposed technique.

\subsection{\textit{\textbf{BdSLImset}} (\textit{\textbf{B}angla\textbf{d}eshi \textbf{S}ign \textbf{L}anguage \textbf{Im}age Data\textbf{set}})}
              After the investigations on existing works mentioned in the previous section, we found lacking of a proper dataset to integrate a neural network architecture. As we want to develop a real time system, our model must be enriched to train such robust classifier, training images should have variations in signs of letter, in backgrounds and lighting conditions. Therefore, we kept these factors in our considerations while collecting training images for this dataset. In our dataset, some images contain desired gesture of letter which is partially obscured, overlapped with something else, or only halfway in the picture. Each image size is less than $200kb$ and the resolution is not more than $700\times 1280$. Fig. \ref{fig4} shows some sample dataset images. Currently, our \textit{BdSLImset} dataset has $10$ different labels sign letters. We collected about $100$ pictures of each gesture. For each letter about $100$ sign images of $10$ persons of different ages and genders have been captured with variety of backgrounds. The dataset is divided into training set and testing set with the ratio of $8$:$2$. After gathering images, we selected the region of each of the hand gestures with a bounding box and labeled them (see Fig. \ref{fig5}). Thus the initial training data is prepared. A comparison between ours and previous dataset has been shown in TABLE \ref{tab2}.

\begin{table}[]
\centering
\caption{Comparisons between datasets used in previous work and in our method.}
\begin{tabular}{@{}ccccc@{}}
\hline
\textbf{\begin{tabular}[c]{@{}c@{}}Related Works\\ \end{tabular}} & \textbf{\begin{tabular}[c]{@{}c@{}}Background\\ \& Lighting \end{tabular}}  &  \textbf{\begin{tabular}[c]{@{}c@{}}Image per\\Class $\times$ \\ Total Classes \end{tabular}} & \textbf{\begin{tabular}[c]{@{}c@{}}No. of\\Signers\end{tabular}}\\ \hline
\begin{tabular}[c]{@{}c@{}}Rahman et\\ al.\cite{b3}\end{tabular} & \begin{tabular}[c]{@{}c@{}}Static\\\end{tabular} & \begin{tabular}[c]{@{}c@{}}$36\times10$\end{tabular} & $10$ \\ \hline
\begin{tabular}[c]{@{}c@{}}Rahman et\\ al.\cite{b18}\end{tabular} & \begin{tabular}[c]{@{}c@{}}Static\end{tabular} & \begin{tabular}[c]{@{}c@{}}$10\times10$\end{tabular} & $10$ \\ \hline
\begin{tabular}[c]{@{}c@{}}Ahmed et\\ al. \cite{b9}\end{tabular} & Static &  \begin{tabular}[c]{@{}c@{}}$37\times14$\end{tabular} & $3$\\ \hline
\begin{tabular}[c]{@{}c@{}}Yasir et\\ al. \cite{b1}\end{tabular} & \textit{N/A} &  \textit{N/A} & \textit{N/A} \\ \hline

\begin{tabular}[c]{@{}c@{}}Our\\(BdSLImset)\end{tabular} & \begin{tabular}[c]{@{}c@{}}Randomized\end{tabular} &  \begin{tabular}[c]{@{}c@{}}$10\times10$\end{tabular} & $10$ \\ \hline

\end{tabular}
\label{tab2}
\end{table}
\subsection{Pipeline of the Proposed Technique}

Our proposed methodology follows a certain pipeline. At first, our training data is fed into the convolutional neural network as mentioned earlier in the overview of Faster R-CNN and eventually makes a feature map based on given aspect ratio and sizes of the model. Then the feature map goes to the RPN and proposes regions with probability of that region being a sign gesture of a letter. 
Fig. \ref{fig8} shows the entire working process of the network. Here, RoI pooling resizes the feature map into fixed sizes for each proposal in order to classify them into a fixed number of classes.

\begin{figure}[h]
    \centerline{\includegraphics[scale=0.67]{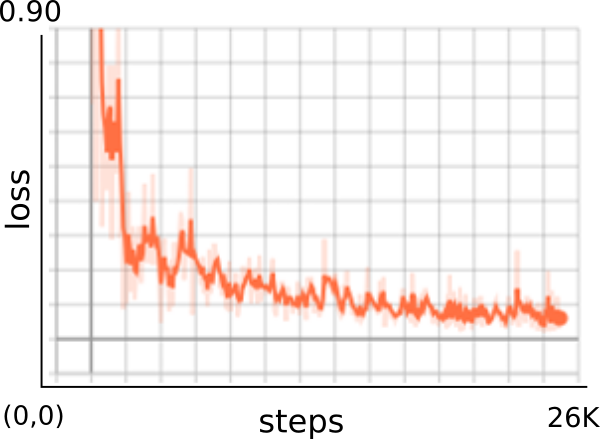}}
    \caption{Loss graph generated from \textit{tensor board}. Initially the loss was about $3.0$ and quickly dropped to below $0.8$ and the training was stop after 26k iterations at loss of 0.075.}
    \label{fig12}
\end{figure}

\section{Experiments and Results}
\label{exp}
        This section represents the experimental results of our proposed BdSL recognition with Faster R-CNN model on the prepared \textit{BdSLImset} dataset. Firstly, we start with experimental setup and then the final outcome and comparisons with other models are presented.
\subsection{Experimental Setup}

            To employ Faster-RCNN model based training in our module, each image were re-sized  and their resolution were kept minimal for training as mentioned earlier. The image set were divided into testing and training set. Training set contains $80$ percent of the images and test set contains $20$ percent.

           Our technique is implemented in \texttt{Tensorflow-GPU V1.5} and \texttt{cuda V9.0}. The training was performed by adopting the \texttt{Faster RCNN Inception V2} model.
           The experiment has been conducted on a machine having CPU from Intel \textregistered. Core\textsuperscript{TM} i7-7500U of $2.7$ GHz, GPU Nvidia $940$mx with $4.00$GB and with $8.00$GB memory on a Windows $10$ operating system.
\subsection{Experimental Result}

\begin{figure*}[!ht]
\centering
\subfloat[][]{\includegraphics[scale=0.082]{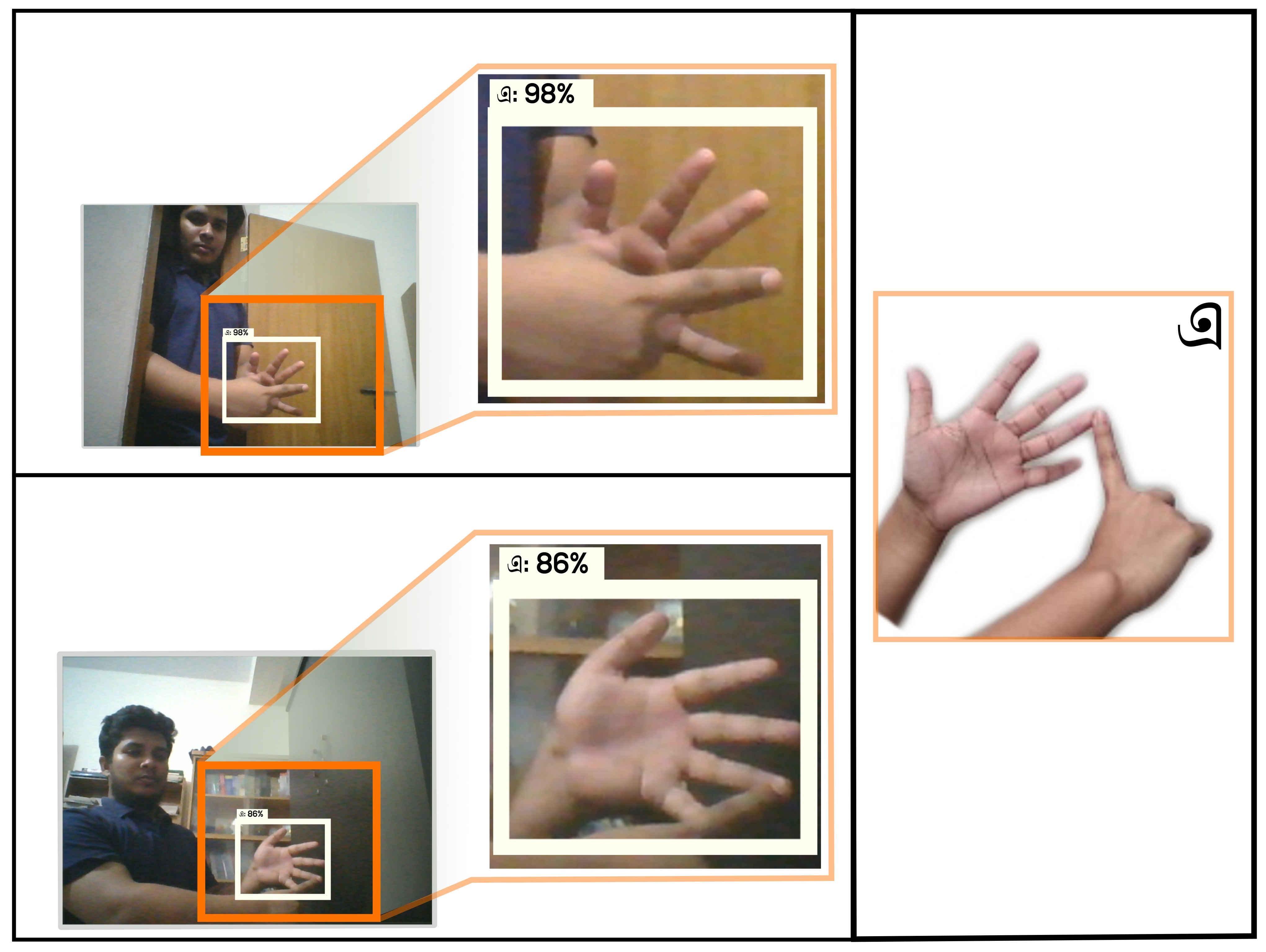}} \hspace{1mm}
\subfloat[][]{\includegraphics[scale=0.082]{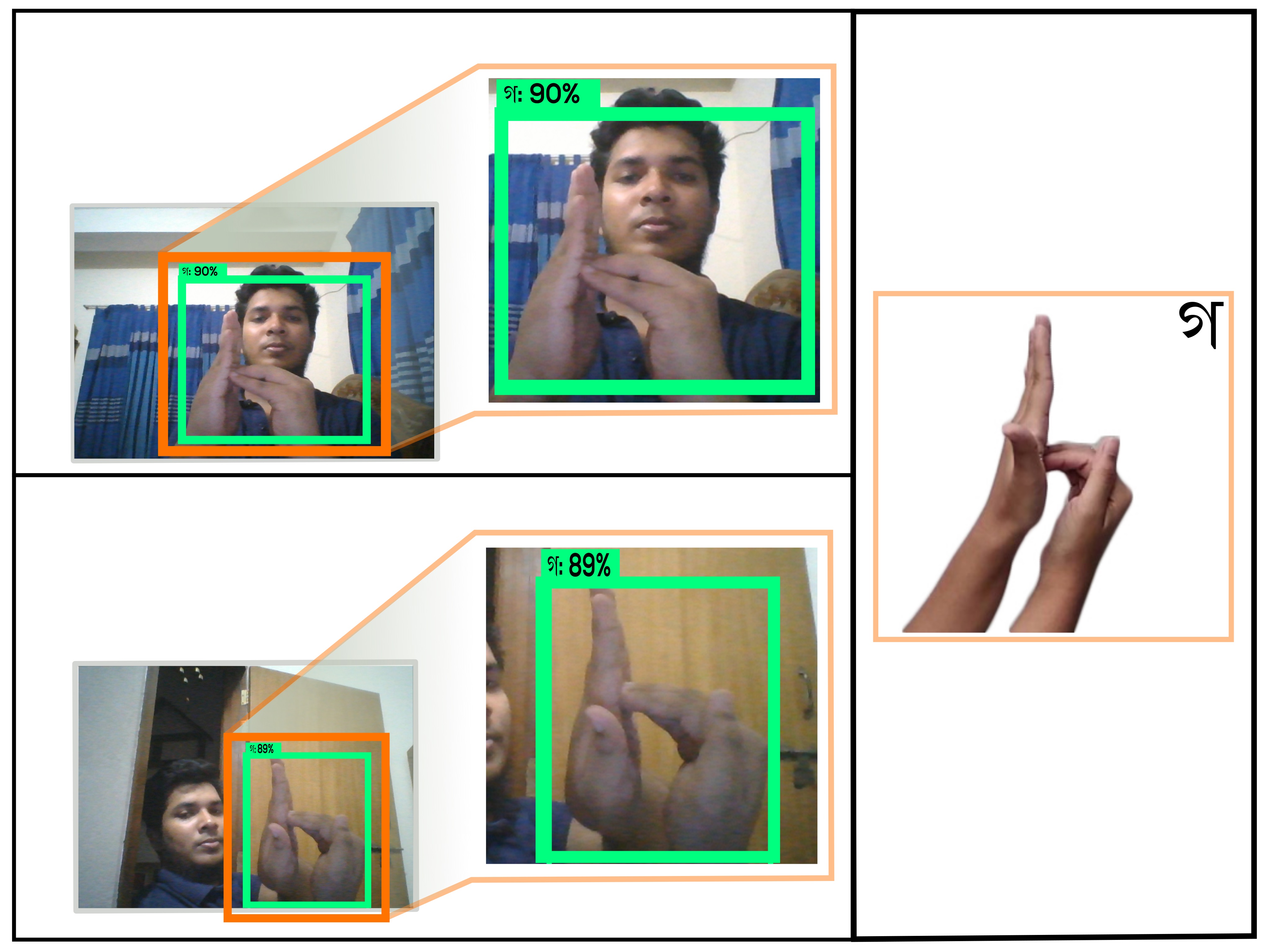}} \hspace{1mm}
\subfloat[][]{\includegraphics[scale=0.082]{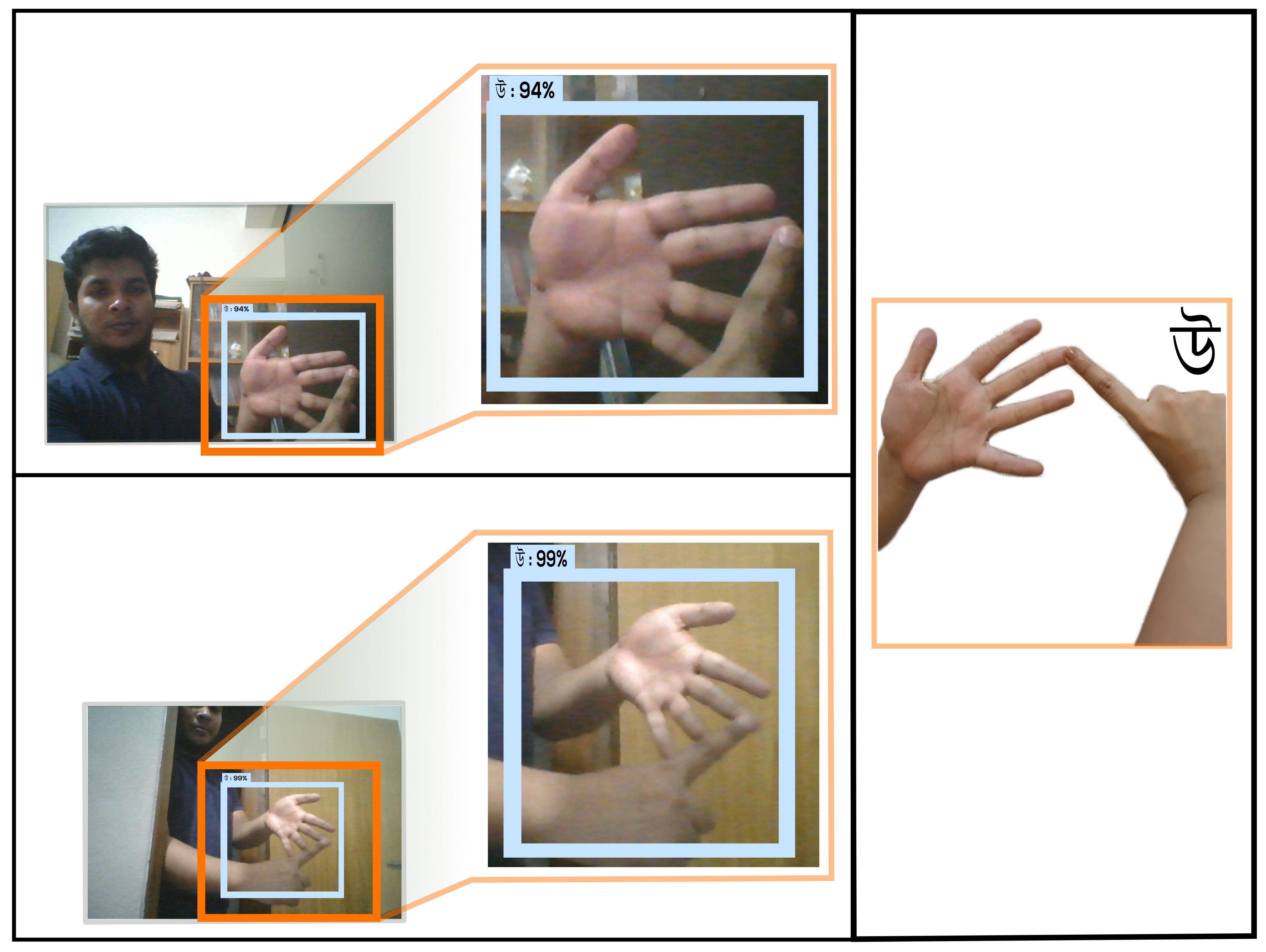}} \hspace{1mm}
\subfloat[][]{\includegraphics[scale=0.082]{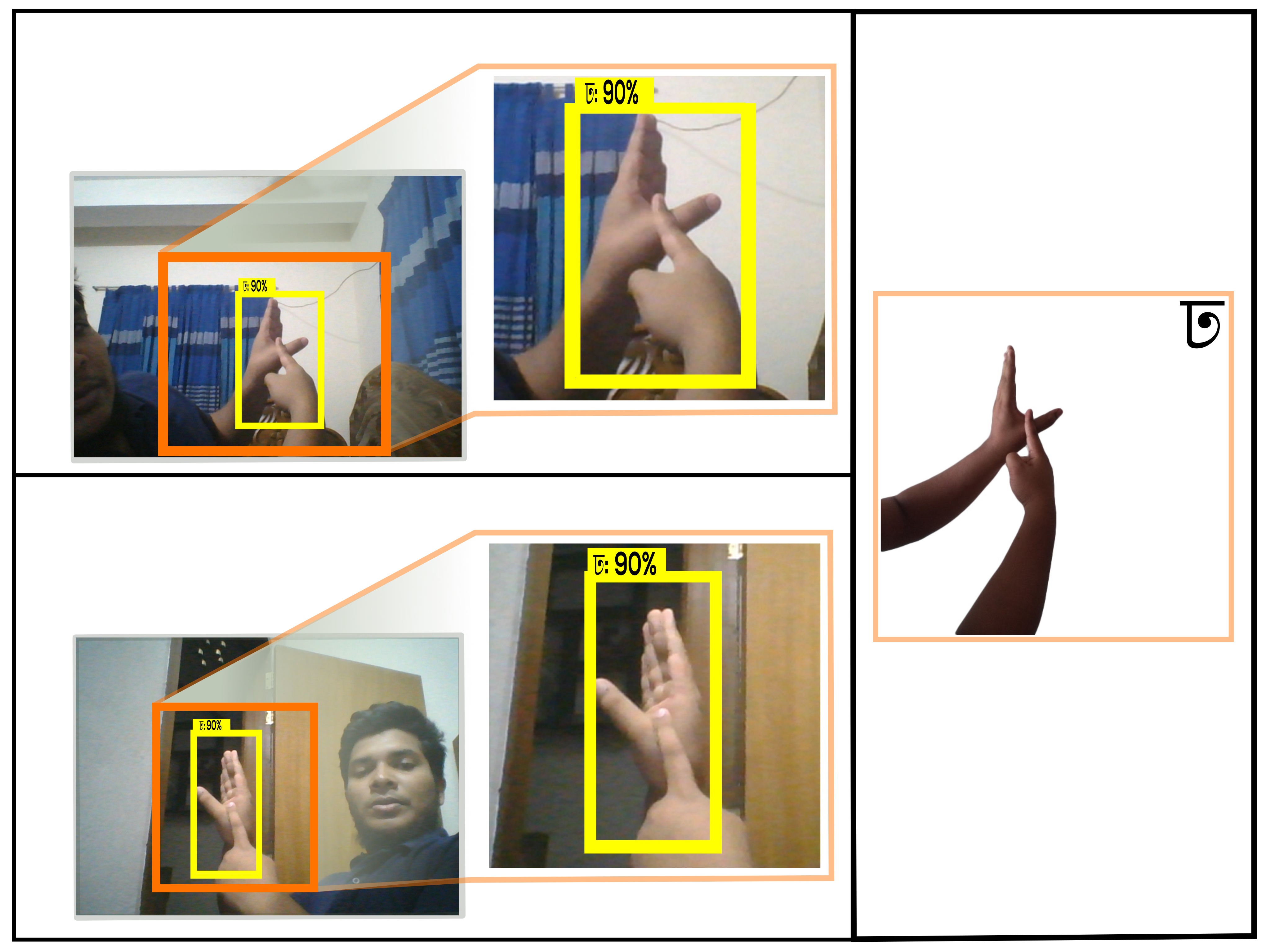}} \hspace{1mm}
\caption{Some results of real time detection `\protect\includegraphics[scale=0.007]{Figure/a.jpg}',`\protect\includegraphics[scale=0.007]{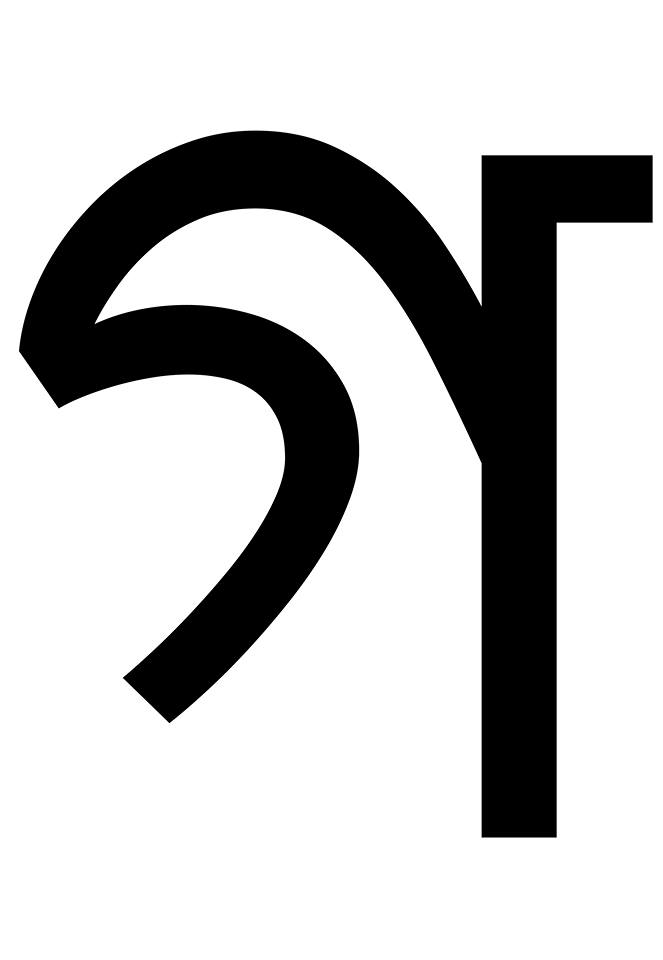}',`\protect\includegraphics[scale=0.007]{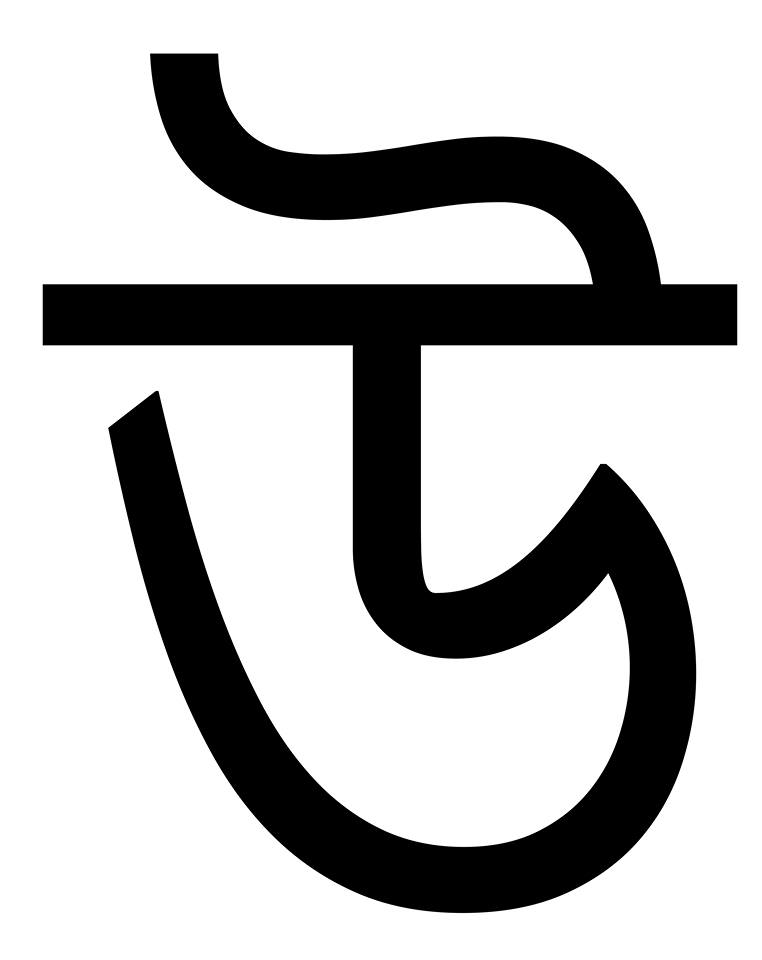}' \& `\protect\includegraphics[scale=0.0056]{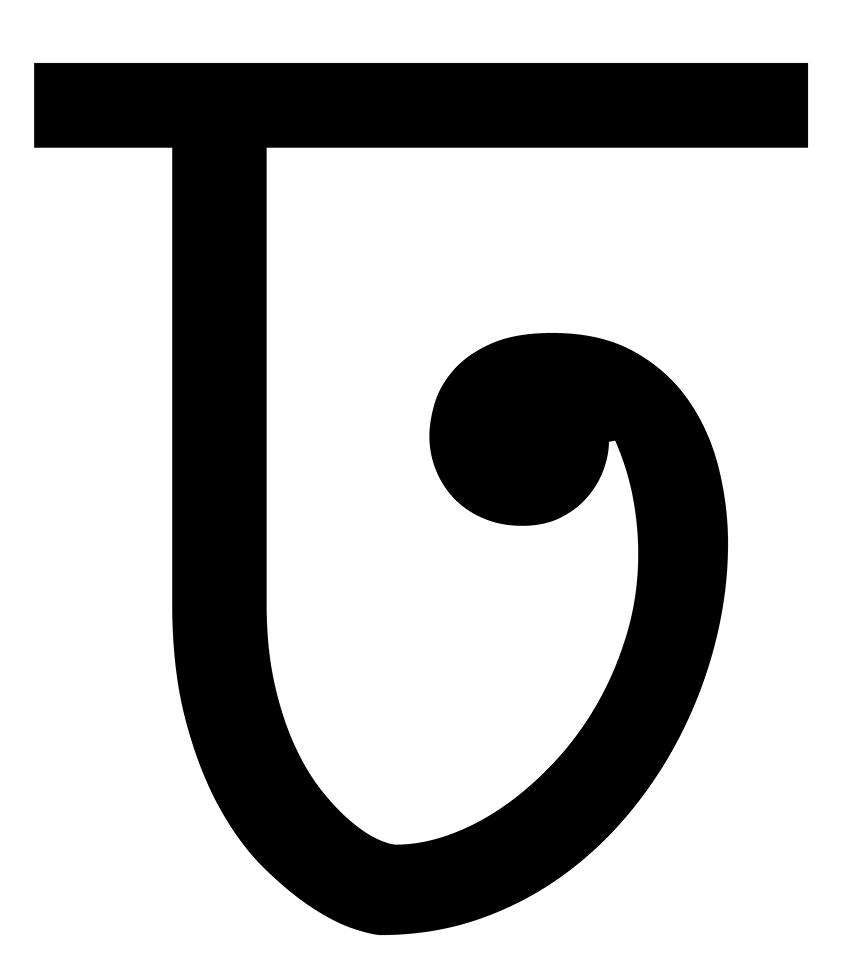}' using our proposed technique. This experiment had been done on various situation, i.e. different background, illumination and angles. In (a), `\protect\includegraphics[scale=0.007]{Figure/a.jpg}' has detection confidence rates of $98\%$ and $86\%$ (\textit{top} \& \textit{bottom row in left respectively}). For each row, $1$\textsuperscript{st} column shows the detection and the  ($2$\textsuperscript{nd} column) has a zoom version of detected portion for better investigation. A reference sign is included at the rightmost column as ground truth collected from \textit{BdSLImSet}. Similarly, (b) shows that  `\protect\includegraphics[scale=0.008]{Figure/ga.jpg}' has detection confidence rates of $90\%$ and $89\%$ (\textit{top} \& \textit{bottom}). In (c) and (d), `\protect\includegraphics[scale=0.008]{Figure/u.jpg}' has  $94\%$ \& $99\%$ (\textit{top} \& \textit{bottom}), and `\protect\includegraphics[scale=0.007]{Figure/dho.jpg}' has  $90\%$ \& $90\%$ confidence rate (\textit{top} \& \textit{bottom}) respectively.}
\label{fig9}
\end{figure*}

            Training was completed with loss of $0.07538$ and accuracy was about $98.2$ percent in average. The detection time was really fast of about 90.03 milliseconds. The training included about $100$ images for each label. As sign letter gesture have similarities among them and take wide range of area of hand and may have different background, sometimes it becomes difficult for the model to differentiate between background and foreground of the image. The experimental process had been conducted for different background and persons, and the recognition time was stable for every test but confidence rate varies in different lighting conditions. Fig. \ref{fig9} shows some result of our testing process.

      Faster R-CNN model takes hours to train the sets of image data. We stopped the training when the loss rate became almost constant to $0.07538$. It took almost $26$K iterations to decrease loss to this rate (see Fig \ref{fig12}). Each step of training reports the loss. It started high and gets lower and lower as training progresses. For our training model, it started at about $3.0$ and quickly dropped below $0.9$.

            Table \ref{tab1} presents the comparisons between proposed and other BdSL detection methods. This comparison is based on their individual dataset and experiment. It is obvious from the table that our method is efficient than the others while considering the performance in real time as well as the accuracy rate.

\begin{figure}[h]
    \centerline{\includegraphics[scale=0.08]{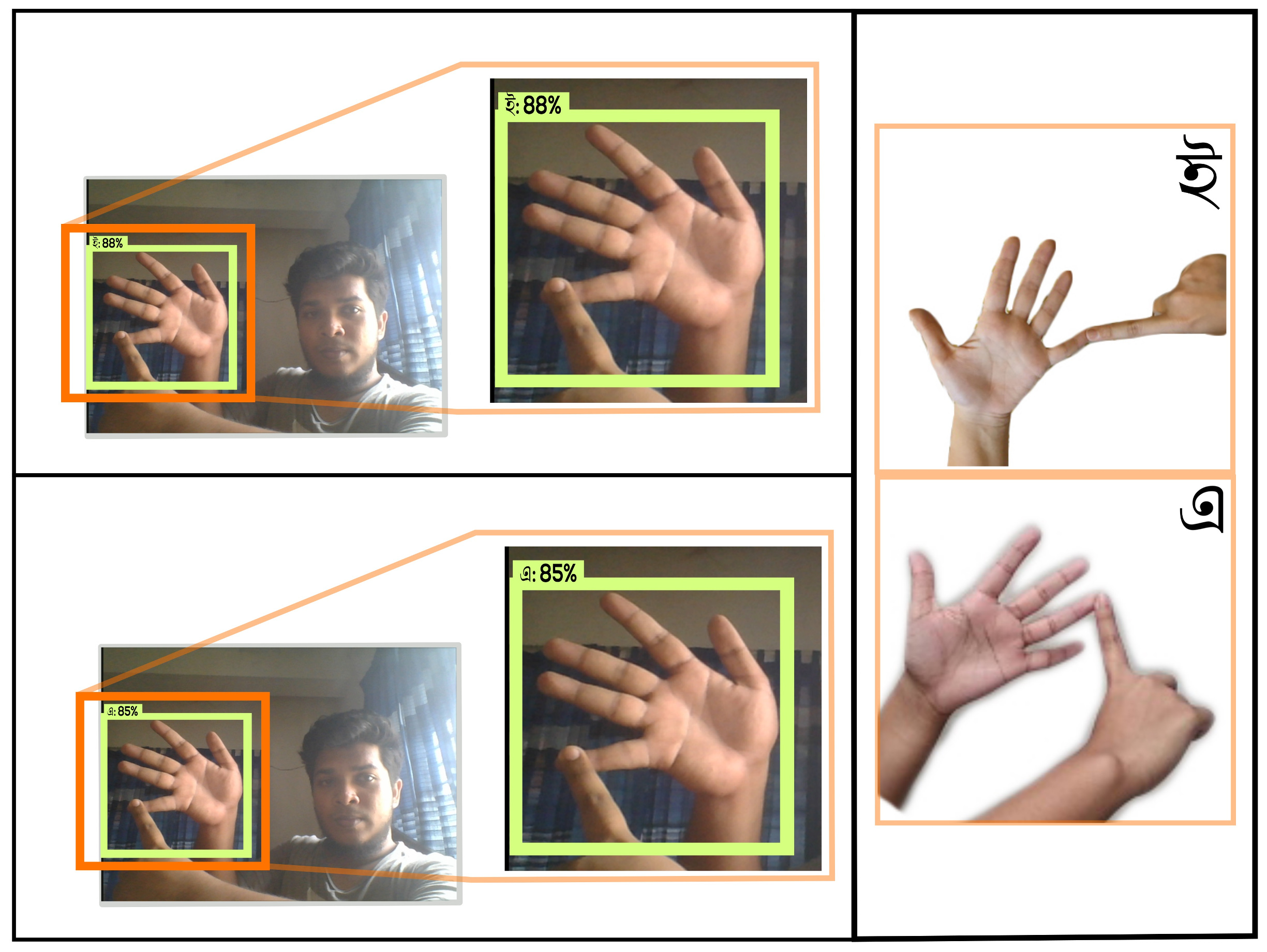}}
    \caption{False recognition
    because of the similarity in the gestures of '\protect\includegraphics[scale=0.008]{Figure/a.jpg}' and '\protect\includegraphics[scale=0.008]{Figure/e.jpg}'. Here, for both rows, the $1$\textsuperscript{st} column shows the detection and the  $2$\textsuperscript{nd} column has a zoom version of detected portion for better investigation. Both the rows show the same signs of '\protect\includegraphics[scale=0.008]{Figure/e.jpg}' while our approach performs perfectly in once case (\textit{top row in left}), while fails in another situation (\textit{bottom row}). A reference sign is included at the rightmost columns as ground truth.}
    \label{fig11}
\end{figure}

            During the testing process, one letter  `{\includegraphics[scale=0.008]{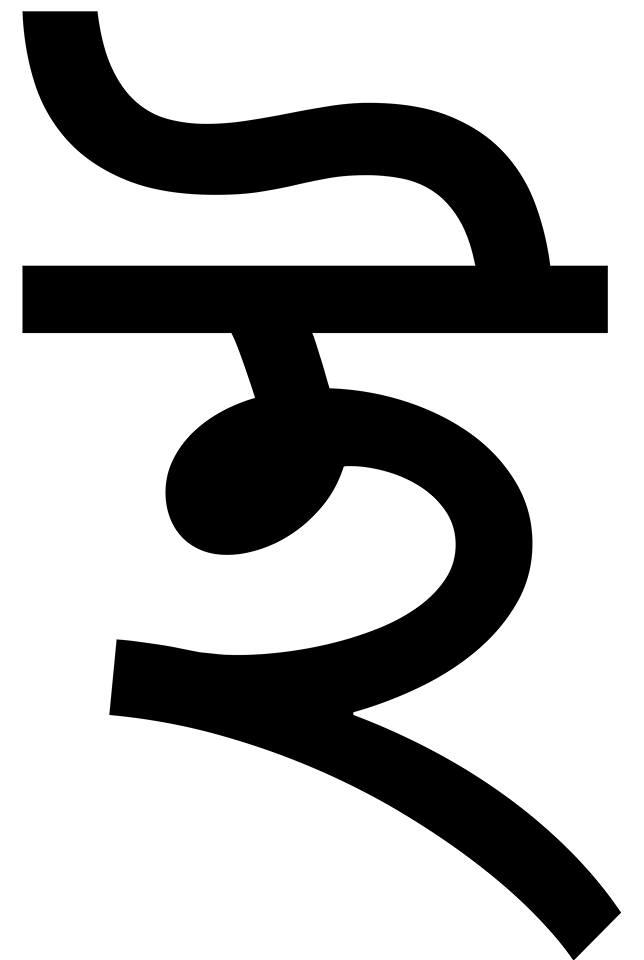}}' was taking more time to be recognized and was giving faulty outcome. The reasons can be due to extreme variation with the background and illumination. The sign letters have much similarities among them which make it complicated to differentiate between these letters. Such as, `{\includegraphics[scale=0.008]{Figure/e.jpg}}' having very similar gesture as `{\includegraphics[scale=0.008]{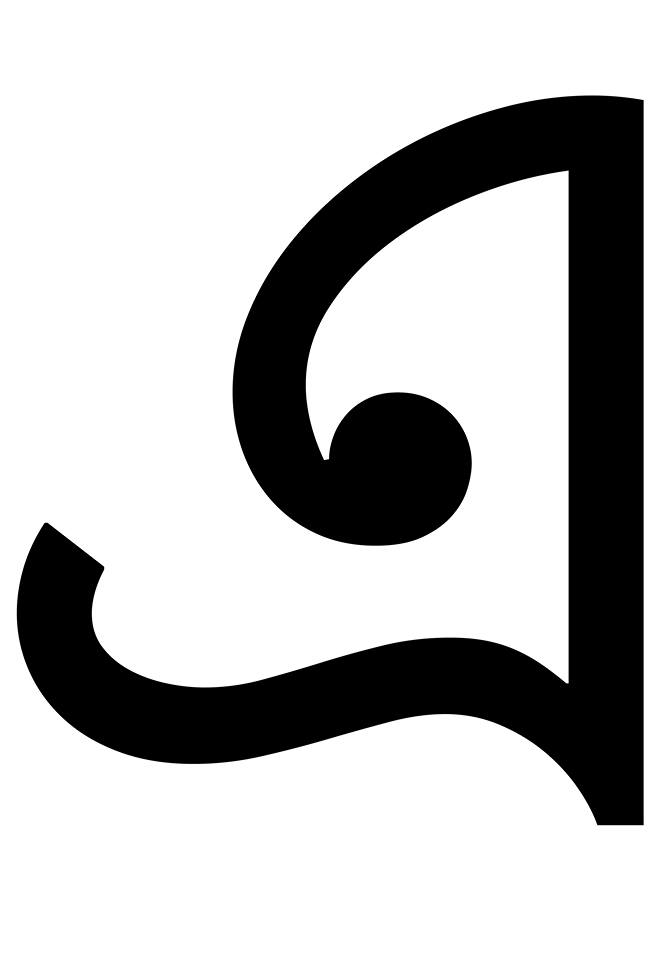}}' which may lead the classifier to a wrong decision. In order to overcome these issues, number of samples in our dataset needs to be enriched more. Fig. \ref{fig11} shows some failed situations during the recognition of `{\includegraphics[scale=0.008]{Figure/e.jpg}}'.

\begin{table}[ht]
\centering
\caption{Comparisons between our method and other works}
\begin{tabular}{@{}cccc@{}}
\toprule
\multicolumn{1}{c}{\textbf{\begin{tabular}[c]{@{}c@{}}Related Works\end{tabular}}} & \textbf{Methodology} & \multicolumn{1}{c}{\textbf{\begin{tabular}[c]{@{}c@{}}Recognition Time\end{tabular}}} & \textbf{Accuracy}  \\ \midrule
\begin{tabular}[c]{@{}c@{}}Rahman et\\ al.\cite{b3}\end{tabular}                            & \begin{tabular}[c]{@{}c@{}}Haar-like\\feature\end{tabular} & \begin{tabular}[c]{@{}c@{}}Real Time\\ (93.55 ms)\end{tabular}                & $96.46\%$                                \\\midrule
\begin{tabular}[c]{@{}c@{}}Ahmed and\\ Akhand \cite{b9}\end{tabular}                           & ANN                                                               & \begin{tabular}[c]{@{}c@{}}Not Real\\ Time\end{tabular}        & $98.99\%$                                \\\midrule
\begin{tabular}[c]{@{}c@{}}Yasir et\\ al. \cite{b1}\end{tabular}                            & CNN                                                               & \begin{tabular}[c]{@{}c@{}}Not Real\\ Time\end{tabular}                   & $97.00\%$                                \\\midrule
\begin{tabular}[c]{@{}c@{}}Rahman et\\ al.\cite{b18} \end{tabular}                       & \begin{tabular}[c]{@{}c@{}}Contour\\Matching\end{tabular}            & Real Time                            & $95.80\%$  \\\midrule
\begin{tabular}[c]{@{}c@{}}Proposed \\ Methodology \end{tabular}                      & \begin{tabular}[c]{@{}c@{}}Faster\\R-CNN\end{tabular}            & \begin{tabular}[c]{@{}c@{}}Real Time\\ (90.03 ms)\end{tabular}                            & $98.20\%$
\\\bottomrule
\end{tabular}
\label{tab1}
\end{table}

\section{Conclusion and Future Work}
\label{conc}

In this paper, we have developed a system that would recognize Bangla Sign Letters in real time. Images of different BdSL signs from our \textit{BdSLImset} dataset were trained by Faster R-CNN based model to solve the problem of sign language recognition. We obtained average accuracy rate of $98.2$ percent and recognition time was $90.03$ milliseconds. Different possible avenues of future exploration of our research are discussed below.

\begin{itemize}
\item Our system has limitations while recognizing the letters, which have many similarities among their pattern. The problem might be overcome with more image data for those letters.
\item Also the image size is a factor, as data training requires huge amount of time. Our research is still ongoing progress. We are collecting more data and training the system to recognize the patterns better.
\item In future, our plan is to evaluate our model by genuine users to sort out its limitations and improve the system. This will also help us see how the system reacts on real life situation and how clearly it can recognize the pattern and interpret effectively.
\item One of our targets is to make a system that recognizes a sequence of signs, concatenate them and translate it into a phrase. In that case, not only a single frame will be considered, but a series of actions, providing a meaningful word or phrase. For this purpose, we are planning to apply recurrent neural network \cite{b16} in future. This attempt will also require our dataset to be refined and processed further.
\end{itemize}




\bibliography{main}{}
\bibliographystyle{IEEEtran}


\end{document}